# Pyramid Medical Transformer for Medical Image Segmentation

Zhuangzhuang Zhang, Weixiong Zhang

Washington University in St. Louis, MO, USA

**Abstract**: Deep neural networks have been a prevailing technique in the field of medical image processing. However, the most popular convolutional neural networks (CNNs) based methods for medical image segmentation are imperfect because they model long-range dependencies by stacking layers or enlarging filters. Transformers and the self-attention mechanism are recently proposed to effectively learn long-range dependencies by modeling all pairs of word-to-word attention regardless of their positions. The idea has also been extended to the computer vision field by creating and treating image patches as embeddings. Considering the computation complexity for whole image self-attention, current transformer-based models settle for a rigid partitioning scheme that potentially loses informative relations. Besides, current medical transformers model global context on full resolution images, leading to unnecessary computation costs. To address these issues, we developed a novel method to integrate multi-scale attention and CNN feature extraction using a pyramidal network architecture, namely Pyramid Medical Transformer (PMTrans). The PMTrans captured multi-range relations by working on multi-resolution images. An adaptive partitioning scheme was implemented to retain informative relations and to access different receptive fields efficiently. Experimental results on three medical image datasets (gland segmentation, MoNuSeg, and HECKTOR datasets) showed that PMTrans outperformed the latest CNN-based and transformer-based models for medical image segmentation.

## 1 Introduction

Accurate and robust medical image segmentation is essential for computer-aided diagnosis, treatment planning, and image-guided surgery [1]. In radiotherapy treatment planning, accurate delineation of tumors can maximize target coverage while minimizing toxicities to the surrounding organs at risk (OARs) [2]. Manual contouring in clinical practice is labor-intensive and time-consuming, thus automated and reliable medical image segmentation has been a research focus for many years.

Accurate segmentation can be achieved by apprehending both global context and local details. Medical image segmentation can be formulated as a problem with three objectives: 1) determining the existence, 2) identifying the shape and rough areas, and 3) refining the boundaries of the tumors or organs of interest. These three objectives require different, from low to high, levels of pixel-wise details on the image. Object localization does not require pixel-wise details of the original image [3], and it is inefficient to compute global attention [1] across the whole image. However, current medical transformers do not explicitly model global context [4, 5] or model global context on full resolution images, bearing unnecessarily high computation costs [1].

Moreover, organs, tumors, and other objects to segment in medical images normally have irregular shapes and different sizes [6]. The current medical transformers do not robustly cater to objects at multiple scales [7] because they arbitrarily divide images. For example, TransFuse [5] uses patch-wise attention, and the patches predetermine fixed receptive fields on the original image. Similarly, Medical Transformer [1] divides an image into 16 pieces for processing. It computes pixel-wise axial attention [1] within a piece, and each piece also has a fixed receptive field on the image; doing so may even separate the same object into different pieces, making segmentation difficult.

Furthermore, arbitrarily dividing images does not consider objects of different shapes and sizes. Intuitively, the position and size of an object inherently determine if it occupies the whole patch (large object), is contained in one patch (small object), or is divided into several patches (an object on cutting boundaries). Neighboring pixels often provide vital information of local details [7], whereas such a dividing scheme may destroy these relations, adversely affecting segmentation outcomes.

In short, while they have been demonstrated to outperform convolutional neural networks (CNNs) based methods, the existing transformer-based methods for medical image segmentation are far from reaching their maximal potential. The primary culprit of their inefficacy is the rigid partitioning an image into patches with fixed sizes, which is doomed to fail to grasp associations of diverse and unknown scales within the image.

We propose a novel, pyramid medical transformer (PMTrans) approach that explores and incorporates multi-resolution attention. To our knowledge, this is the first multi-scale and multi-resolution medial image transformer, which is not only designed for medical images but also applicable to general images such as scenery pictures.

The new method has three prominent features. Firstly, the new model uses a pyramid architecture with four branches: three transformer branches and one CNN branch. The three transformers work on three different image scales to capture different ranges of correlations within the image, i.e., short-range, mid-range, and long-range associations. We use down-sampled images (half and 1/4 sizes of the original) to capture mid-range and long-range relations and use the original image to capture local, finer details. This pyramid design not only improves the efficiency of global context modeling but also accommodates objects with diverse sizes and shapes robustly.

Secondly, we abandon the existing partitioning schemes [1, 4, 5, 8] and adopt gated axial attention and CNN as building blocks. We introduce local gated axial attention layers [1] in the short-range transformer. We only compute gated axial attention within a certain span, which is much shorter than the height or width of the original image. This local gated axial attention module does not concern global attention because pixels only attend to their neighbors. We focus on local finer details in this short-range transformer, and the global context is ignored for efficiency. We use global gated axial attention layers on downsampled input images in the mid-range and long-range branches to capture global attention effectively. Since global axial attention may be relatively costly to compute, we apply it to down-sampled images, which means we cover larger receptive fields on the original image at low costs.

Thirdly, we design and implement a new fusion and label guidance scheme to integrate feature maps of different scales from the three multi-scale transformers and image-specific features from the CNN branch. We preserve low-level context without deep CNNs and retain the feature extraction characteristics of CNN and transformer models [5]. As for label guidance, we inject deep supervision [2] into different feature map scales in the training process, which stabilizes the training and alleviates a potential gradient vanishing problem [9].

We tested the performance of the proposed PMTrans model on three datasets: gland segmentation (microscopic) dataset [10], MoNuSeg (microscopic) dataset [11], and HECKTOR (PET/CT) dataset [12]. The results showed that the new model outperformed the latest CNN-based and transformer-based models.

## 2 Related Work

Before deep learning was introduced to the medical field, most automated medical image segmentation models are atlas-based [13, 14], statistical-based [15], and shape-based [16]. With the great successes of CNNs, particularly on images, many CNN-based medical image segmentation models have been developed. The most popular is the encoder-decoder architecture proposed in the U-net [17], subsequently improved and extended. Some architectures replace the vanilla stacking convolutional layers with other backbones, resulting in, e.g., residual U-net [18] and dense U-net [19] as well as their implementation innovations, such as U-net++ [20], V-net [21], Y-net [22]. These CNN-based methods work on various modalities, target locations, and segmented targets. They not only perform well on a wide range of images but also expand the capacity of convolutional neural networks.

Although most latest computer vision models use CNNs as building blocks, the drawback of CNNs is also noticeable and profound. Convolutional layers are not designed to capture long-range dependencies because they aggregate local information within their filter regions across one layer to the next; capturing long-range associations requires deep networks or very large filters, resulting in models with a large number of parameters that are difficult and costly to train. A few approaches have been proposed to address this issue: 1) using the atrous convolution [7] so that the receptive fields enlarge without increasing the number of parameters; 2) incorporating statistical modeling like Markov Random Field to support global dependencies [6]; 3) using a feature pyramid structure to deal with multiple image resolutions [23].

Transformer [24] is a ground-breaking deep-learning technique for self-attention. It makes the self-attention mechanism practically feasible at a global scale so that long-range dependencies can be learned efficiently. Transformer-based models revolutionize the natural language processing (NLP) field, reminiscent of CNNs revolutionizing the field of computer vision (CV). Effective transformer models like BERT [25] and GPT [26] outperform the previous sequence-to-sequence NLP models in many tasks like machine translation and text generation.

The idea of the transformer has been attempted in developing attention-based CV models. One major difficulty of adopting transformers for CV tasks is the high computational complexity and cost for modeling and learning global attention. In NLP tasks, sentences are normally less than 500 words, which is manageable for computing all pairwise attention of the words in a sentence. However, for an $H$ by $W$ image, $HW$ pixels give rise to a quadratic order of (i.e., $H^2W^2$) pairs of attention weights, e.g., more than 68 billion pairs of attention for a 512 by 512 image. This direct adaptation of the idea for images is not scalable. To address this problem, Vision Transformer (ViT) [24] partitions a 2D image into patches and then converts the patches with a positional encoding to a 1D sequence to feed into transformer blocks. Instead of computing pixel-to-pixel attention, it computes patch-to-patch attention. For an $H$ by $W$ image with a $p \times p$ patch size, this scheme reduces to $HW/p^2$ patches and $H^2W^2/p^4$ pairs of attention weights. i.e., roughly one million pairs of attention for a 512 by 512 image with a patch size of 16 by 16. Segmentation Transformer [8] replaces the encoder in a CNN-based method with a transformer with superior performance for end-to-end segmentation tasks.

Beyond using patches to reduce the number of parameters of self-attention, approximation alternatives have been considered to avoid the quadratic computational cost. Learning global self-attention amounts to computing all pairwise relations. Nevertheless, full attention is not the only solution. Axial-Deeplab [27] introduces an approximation method that substitutes vanilla global attention with axial attention by computing the vertical and horizontal axial attention, which reduces the complexity from $O(H^2W^2)$ to $O(HWm)$ where m is the axial span of the attention operation. Sparse transformer [28] extends this approximation method and generates long sequences to reduce the complexity from $O(H^2W^2)$ to $O(HW\sqrt{HW})$.

Besides high computation cost, the second difficulty that vision transformers encounter is that they require colossal datasets for pre-training. Randomly initiated model weights let each patch attend to any other patch randomly at the start of the training, so a large amount of training samples is needed to learn where to "look" and aggregate information to pass onto the next layer. Compared to CNN-based models, where filters slide through the image stride by stride, the self-attention mechanism offers more liberty to attend yet facing higher learning costs. Data-efficient Transformer (DeiT) [29] addresses this issue with a distillation mechanism, making the model trainable on mid-sized datasets.

In the medical image segmentation field, transformer-based methods have not yet been fully explored. High-performance transformers for medical images face two bottlenecks: the computational cost and lack of sufficient training data. In clinical practice, high computational cost posts a burden on the hardware requirements and increases the patient's wait time which may not be acceptable. Moreover, medical datasets are typically much smaller than daily images. ViT [24] uses more than 300M regular images to pre-train the model. Nevertheless, none of the currently available medical datasets has a sufficient amount of training images because medical data collection and annotation demand medical expertise [2].

The current medical image segmentation transformers fall into two categories. The first integrates the attention mechanism into their CNN models as a performance-boosting component, as done in Attention U-net [30] and volumetric attention model [31]. The second type of model is mainly built with transformer layers and uses CNNs as feature extraction blocks. For example, TransUnet [4] inserts attention layers at the end of the CNN-based feature extraction encoder, and the model relies on the pre-trained weights of ViT. Medical Transformer[1] uses gated axial attention and a local-global (LoGo) strategy, where the gated axial attention module is to address the lack-of-data issue, and the LoGo strategy reduces the computation cost of global self-attention. TransFuse [5] combines feature maps from a CNN and a transformer. However, the rigid patching scheme hinders the performance of all previous medical transformers, but without it they would suffer from drastically increased computation cost. Our approach effectively avoids the patching strategy, and wisely applies small axial attention window at multiple scales, capturing dependencies of all ranges efficiently.

## 3 Pyramid Medical Transformer (PMTrans)

### 3.1 Overview

Our approach to capturing both global and local relations, with a manageable computation cost, is to build a pyramid deep-learning architecture, which we call the pyramid medical transformer (PMTrans, Fig. 1). The new method has four branches: three transformer branches are built with gated axial attention blocks [1], and the fourth branch with CNN blocks.

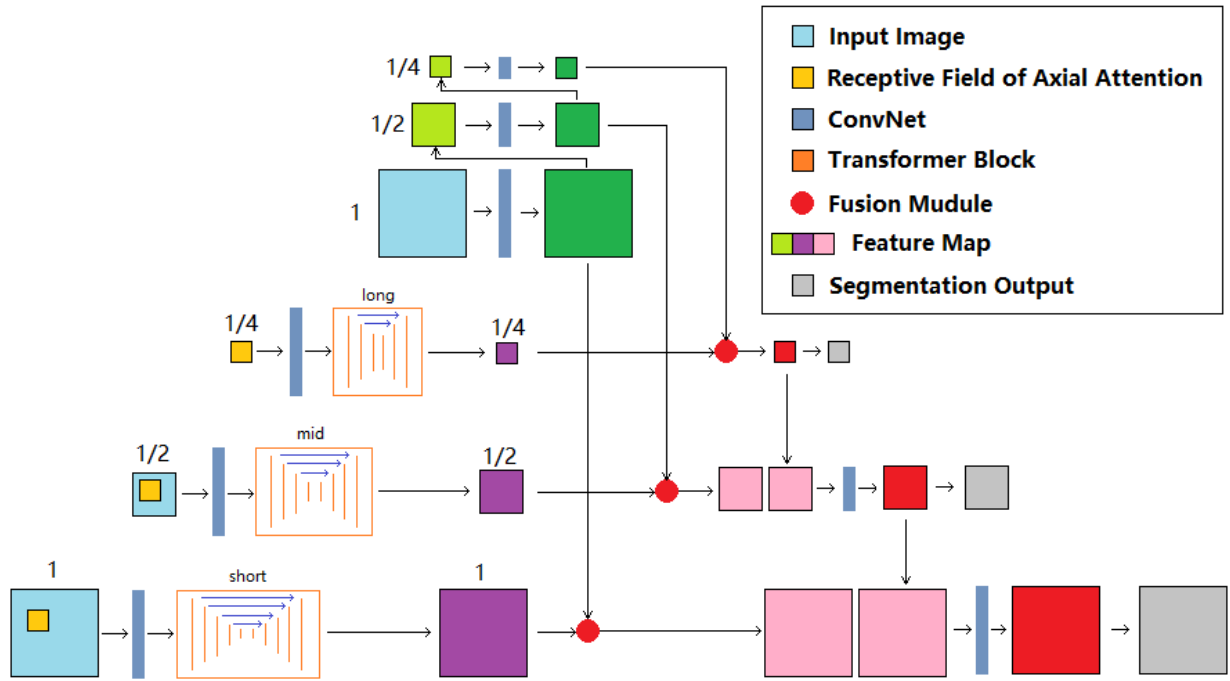

Fig. 1 Structure Overview. The pyramid medical transformer has three transformer branches (short, mid, and long-range branches) and a CNN branch. Input images are rescaled to ½ and ¼ for mid-range and long-range branches, respectively. Feature maps from the transformer branches (in purple) are fused with those feature maps from the CNN branches (in green) at different scales. Deep supervisions are injected at ½ and ¼ scales during the training phase.

The three transformer branches (Fig. 1) are designed to capture short-range, mid-range, and long-range associations among the pixels of an image. They have similar structures but take three different scales of the image ($Height \times Width \times Channel$) as input, which are $H \times W \times C$, $\frac{H}{2} \times \frac{W}{2} \times C$, and $\frac{H}{4} \times \frac{W}{4} \times C$, respectively. They all have an initial convolution block for basic feature extraction at the beginning. This convolution block has three convolution layers followed by batch normalization and the ReLU function [1]. They then have an encoder-decoder-style transformer block to compute self-attention among the pixels. These transformer blocks are built with gated axial attention layers [1] and have skip connections between the encoder and decoder paths. The short-range transformer block has a depth of five, the mid-range block of four, and the long-range block of three accommodating the different sizes of their input feature maps. The outputs of these three branches have the same height and width as their inputs and are subsequently integrated with the feature maps from the CNN branch. The combined results are then up-sampled to restore the original resolution. Using multi-scale input images improves the segmentation of objects with different sizes and shapes.

We design the fourth branch with CNN-based residual blocks (Fig. 1) for extracting features in the input image. Comparing with CNN-based models, we only use a shallow network for the task. By fusing the feature maps from the transformers and CNN, our model preserves low-level context and global relations. Deep supervision [9] is introduced to the two down-sampled feature maps. We build two auxiliary classification layers at the two down-sampled resolutions and use the down-sampled ground-truth label maps correspondingly to stabilize the training process and alleviate the potential gradient vanishing problem [9].

### 3.2 Gated axial attention

In the three transformer branches, we use the gated axial attention, an extension to axial attention [1]. Axial attention computes approximate self-attention. Consider an input image or a feature map X. The self-attention Y of X is computed as

$$Q = W_q X, \quad K = W_k X, \quad V = W_v X, \tag{1}$$

$$Y = softmax(\frac{QK^T}{\sqrt{d_k}})V, \tag{2}$$

where Q, K, and V are query, key, and value matrices computed by weight $W_q$, $W_k$, and $W_v$, respectively; $d_k$ is the number of dimensions of the keys and queries [32]. Note that dividing $QK^T$ by $\sqrt{d_k}$ is to counteract the effect of having small gradients due to large dot products because of high-dimensional keys and queries [32]. Assume that we have a feature map $X \in R^{H \times W \times C_{in}}$, where $H$, $W$, and $C_{in}$ represent the height, width, and number of channels, respectively. There are $H \times W$ tokens and each of which has $C_{in}$ dimensions. We need to compute the attention between each super-pixel and all tokens on the feature map, i.e., a total of $H^2 W^2$ pairs of attention to compute. Self-attention is effective for capturing relations of any distance. All pairs of attention have equal opportunities to be selected by the model regardless of how far two tokens are apart from each other. However, an obvious drawback of self-attention is the overall computational cost.

As an approximation, axial attention [33] applies self-attention computation along the height and width axes of the image. It first computes the attention between each pixel and all pixels in the same column (i.e., along the height) and then the attention between each pixel and all pixels in the same row (i.e., along the width). Note that the attention along the height axis aggregate information in the same column, and subsequently, the attention along the width axis aggregates the combined information in the same row. When each pixel attends to all pixels in the same row, it also indirectly or approximately attends to all other pixels in the feature map. Doing so reduces the computation from $HW \times HW$ to $HW \times (H + W)$. For an input feature map $X \in R^{H \times W \times C_{in}}$, the updated self-attention mechanism [1] with positional encodings along the width axis is:

$$y_{ij} = \sum_{w=1}^{W} softmax(q_{ij}^T k_{iw} + q_{ij}^T r_{iw}^q + k_{iw}^T r_{iw}^k)(v_{iw} + r_{iw}^v), \tag{3}$$

where $i$ and $j$ are the pixel positions along the width and height axes; and $r^q_{iw}$, $r^k_{iw}$, and $r^v_{iw}$ are relative position encodings regarding queries, keys, and values, respectively. Note that the relative position embedding is also along the width axis. The axial attention along the height axis is similar.

Gated axial attention [1] adds weights (i.e., gates) to the relative position encodings in Eqn. (3). These gates would reduce the effect of these encodings if the model cannot learn accurate position encodings with small medical datasets. The gated axial attention computes

$$y_{ij} = \sum_{w=1}^{W} softmax(q_{ij}^T k_{iw} + G_Q q_{ij}^T r^q_{iw} + G_K k_{iw}^T r^k_{iw})(G_{V1} v_{iw} + G_{V2} r^v_{iw}), \quad (4)$$

where $G_Q$, $G_K$, $G_{V1}$, and $G_{V2}$ are gates that control the influence of position encodings for the queries, keys, and values [1], respectively.

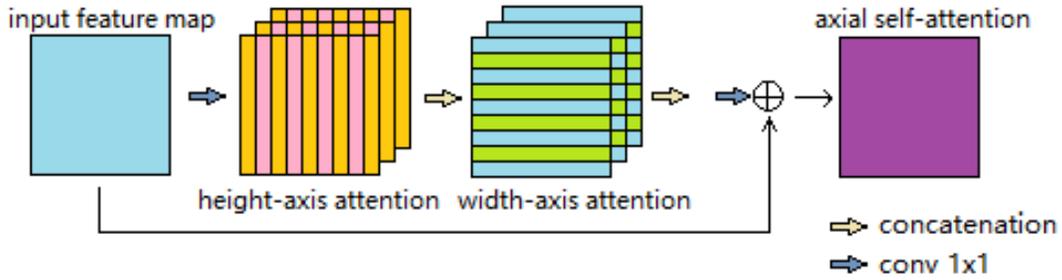

Fig. 2 Encoder block structure. Shown here is global axial attention with a span that covers the whole axis.

Our model builds the encoders of the transformers with gated axial attention layers (Figure 2). Each of the encoder blocks has a $1 \times 1$ convolution layer, two multi-head axial attention layers (one each along the height and width axes), and one 1×1 convolution layer (Fig. 2). Residual connection is also included in encoder blocks. A decoder block consists of a series of one convolution layer, one bilinear up-sampling layer, and one ReLU layer.

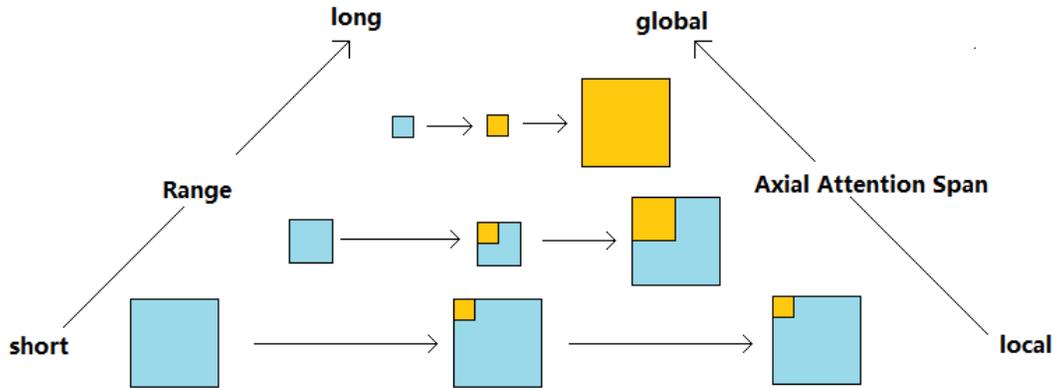

Fig. 3 Pyramid Structure of the Transformer Branches. The axial attention span is the same across the three branches, but they have different receptive fields on the input image.

The input to the short-range branch is the original image ($H \times W \times C$). It uses local gated axial attention blocks to model local context. We set the span of the gated axial attention to $\frac{H}{4}$ and $\frac{W}{4}$. Each pixel attends to a subregion ($\frac{H}{4} \times \frac{W}{4}$) centered on itself (Fig. 3). We avoid the trivial partitioning scheme that would potentially cut off valuable relations. Compared to the patching solution proposed by MedTrans [1], which partitions the original image into $\frac{H}{4} \times \frac{W}{4}$ patches, the gated axial attention that we use has the same complexity, yet our partitioning scheme is adaptive. MedTrans computes global gated axial attention on $\frac{H}{4} \times \frac{W}{4}$ patches, and each pixel attends to $\frac{H}{4} + \frac{W}{4}$ other pixels, giving rise to a total of $\left(\frac{H}{4} + \frac{W}{4}\right) \times HW$ pairs of attention. In our model, we have an adaptive partitioning scheme in which each pixel attend to $\frac{H}{4} + \frac{W}{4}$ other pixels. The receptive field of gated axial attention is $\frac{H}{4} \times \frac{W}{4}$ on the original image. The transformer block in this branch has an encoder-decoder architecture, with five attention layers in the encoder and five in the decoder. Skip connections are implemented to enhance the gradient flow.

The input to the mid-range branch is a down-sampled version of the original to $\frac{H}{2} \times \frac{W}{2} \times C$. We still set the span of the gated axial attention to $\frac{H}{4}$ and $\frac{W}{4}$. However, the receptive field of the gated axial attention is $\frac{H}{2} \times \frac{W}{2}$ on the original image, which means that we can model longer yet still not global relations using this branch. Each pixel attends to $\frac{H}{4} + \frac{W}{4}$ pixels, and there are $\left(\frac{H}{4} + \frac{W}{4}\right) \times HW/4$ pairs of attention in total. We stack four layers in the encoder-decoder architecture of this branch.

For the long-range branch, the input is further down-sampled from the original image to $\frac{H}{4} \times \frac{W}{4} \times C$. Using the same span ($\frac{H}{4}$ and $\frac{W}{4}$) as in the mid-range branch, we can now efficiently model global attention. The receptive field of this gated axial attention is $H \times W$ on the original image, but we only need to compute $\left(\frac{H}{4} + \frac{W}{4}\right) \times HW/16$ pairs of attention. Compared to the global branch of MedTrans [1], which involves $(H + W) \times HW$ pairs of attention, this long-range branch only has 1/216 of the attention pairs used in MedTrans. We stack three layers in this encoder-decoder architecture.

In summary, the three transformer branches of the PMTrans model have different receptive fields on the input image, but all use fixed-range gated axial attention. With this pyramid architecture, we model relations at three different resolutions and ranges efficiently. This design not only effectively caters to objects at different scales but also computes global attention without computing full resolution self-attention. Moreover, we avoid arbitrarily partitioning the image into patches. Instead, we preserve a pixel's local relations by computing axial self-attention within a patch centered on itself.

### 3.3 Fusion and deep supervision

We introduce the fourth, CNN branch to the model to extract features in the input image (Fig. 1). This branch is built with residual convolution blocks, stacking convolution layers with skip connections for gradient flow enhancement [34]. Feature maps from the CNN branch are fused into those from the transformer branches at different scales (Fig. 1), and the pyramid fusion scheme is built with attention gates proposed by the attention U-net [30]. Instead of simply concatenating feature maps and let convolution layers learn to fuse them, we add weights to integrate them consciously with attention-aware learnable parameters at different scales. Using these attention gates at multiple scales enables the model to extract relevant information from both transformers and CNNs, exploiting the feature extraction and information aggregation capabilities of both techniques [35]. We introduce deep supervision by adding auxiliary classifiers after the pyramid fusion of feature maps. Label guidances at different scales are provided at these three branches during training, and only the full resolution output is used for testing and evaluation. These deep supervisions enforce intermediate feature maps to be semantically discriminative at each scale [36].

## 4 Experiments and Results

### 4.1 Datasets

We used three medical image datasets to evaluate the model: gland segmentation (GLAS) [10], MoNuSeg [11], and HECKTOR dataset [12].

- The GLAS dataset contains 85 microscopic images for training and 80 for testing, each with a resolution of 128 by 128 [10].
- The MoNuSeg dataset has 30 microscopic images for training and 14 images for testing, each with a resolution of 1000 by 1000 [11].
- The HECKTOR dataset contains 201 training cases and 53 testing cases, each of which has $144 \times 144 \times 144$ CT/PET volumes [12]. We stacked and fused the CT and PET slices to feed into the network.

We set the training/validation split for 65/20, 25/5, and 180/21 for GLAS, MoNuSeg, and HECKTOR, respectively. We applied random shifting, rotation, and flipping for data augmentation. We performed cross-validation on both datasets to select the best model weights.

Fig. 4 Segmentation Comparison among Residual U-net, Medical Transformer, and Pyramid Medical Transformer. We mark focal regions where models have distinguishable predictions. We cropped 128 by 128 regions from the MoNuSeg and HECKTOR images to better present segmentation details.

|  | Input | Residual U-net [34] | Medical Transformer [1] | Pyramid Medical Transformer (ours) | Ground Truth |
|---|---|---|---|---|---|
| Glas | 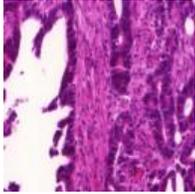 | 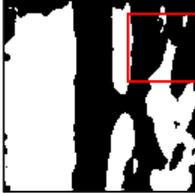 | 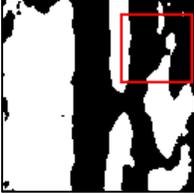 | 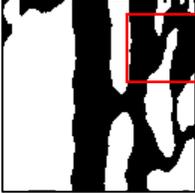 | 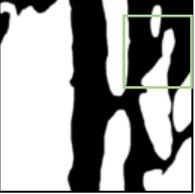 |
| | 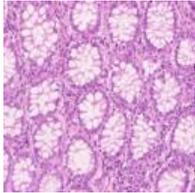 | 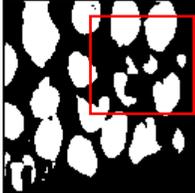 | 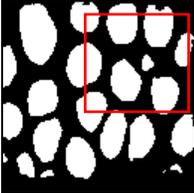 | 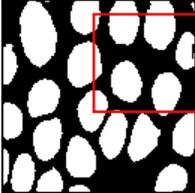 | 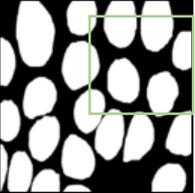 |

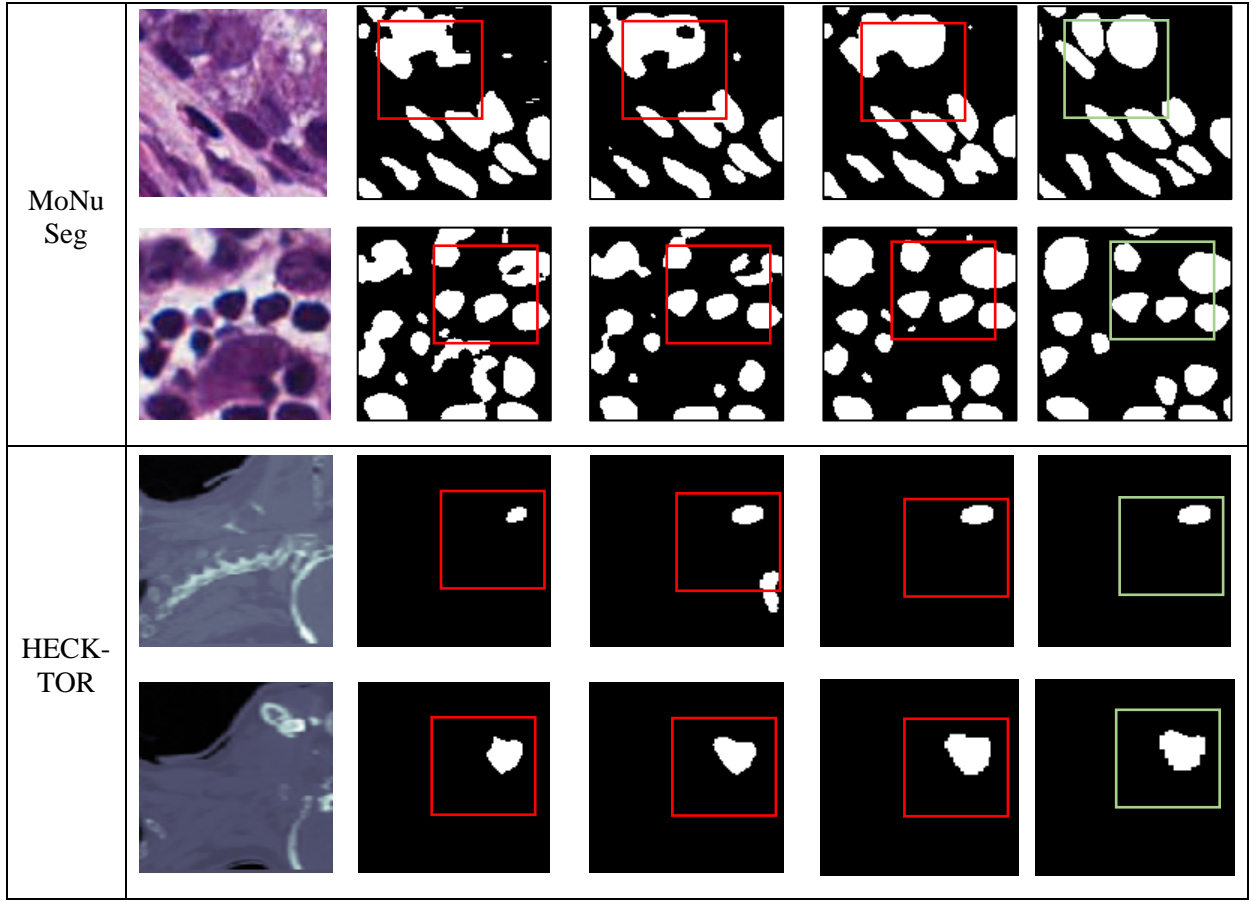

## 4.2 Training Details

The Pyramid Medical Transformer was implemented with Pytorch and trained on an RTX 3090 GPU. We set the batch size to 2 and the learning rate to 0.001 with a polynomial learning rate scheduler. The model was trained using Adam optimizer [37], and the training process finished within 400-500 epochs. The binary cross-entropy loss was defined as

$$\ell_{bce}(\hat{z}, z) = -\sum_{i=1}^{H \times W}[z_i \ln\hat{z}_i + (1 - z_i)\ln(1 - \hat{z}_i)], \quad (5)$$

where $\hat{z}$ is the prediction label map and $z$ is the ground truth label map. As suggested by [1], we did not activate the gates' training in the first 10 epochs. We used the binary cross-entropy loss for training and evaluated the final models by Dice Similarity Coefficient (DSC), defined as

$$DSC = \frac{2|True \cap Pred|}{|True| + |Pred|}, \quad (6)$$

where $|True|$ and $|Pred|$ stand for the numbers of pixels in the ground-truth and predicted masks, and $|True \cap Pred|$ is the number of pixels in the overlapping region [6].

## 4.3 Results

We carried out a quantitative comparison of PMTrans with several state-of-the-art medical image segmentation methods, including both CNN-based methods (FCN [38], U-net [17], U-net++ [20], and Res-Unet [34]) and attention-based methods (Axial Attention U-net [27] and Medical

Transformer [1]). The results are shown in Table 1. Note that all results on the HECKTOR dataset are based on the validation set because we do not have access to the ground truth of the test set.

Our PMTrans method outperformed all six baseline methods on both datasets (Table 1). PMTrans outperformed the best convolution-based baseline by 3.36% on the GLAS dataset, 0.75% on the MoNuSeg dataset, and 1.41% on the HECKTOR dataset. PMTrans surpassed the best transformer-based baseline method (MedTrans) by 0.57%, 0.68%, and 2.21% on the GLAS, MoNuSeg, and HECKTOR datasets, respectively. We also visually compared the segmentation outcomes of these methods using the test images. Figure 4 exhibits some of the results from Residual U-net [34], Medical Transformer [1], and PMTrans. We chose example slices with objects of various sizes and shapes to demonstrate the model performances. Our model catered to (was particularly suitable for) objects at different scales and performed well on objects with various shapes and sizes; the contours from PMTrans closely resembled the ground truth.

| Models | GLAS [10] | MoNuSeg [11] | HECKTOR[12] |
|---|---|---|---|
| **FCN [38]** | $0.64\pm0.06$ <0.01* | $0.32\pm0.06$ <0.01* | $0.75\pm0.10$ <0.01* |
| **U-net [17]** | $0.75\pm0.08$ <0.01* | $0.79\pm0.09$ 0.05* | $0.75\pm0.07$ <0.01* |
| **U-net++ [20]** | $0.77\pm0.07$ <0.01* | $0.79\pm0.09$ 0.04* | $0.79\pm0.05$ <0.01* |
| **Residual U-net [34]** | $0.78\pm0.08$ 0.01* | $0.78\pm0.10$ <0.01* | $0.78\pm0.08$ <0.01* |
| **Axial Attention U-net [27]** | $0.75\pm0.08$ <0.01* | $0.77\pm0.10$ <0.01* | $0.77\pm0.07$ <0.01* |
| **Medical Transformer [1]** | $0.80\pm0.05$ 0.11 | $0.79\pm0.08$ 0.04* | $0.78\pm0.08$ <0.01* |
| **Pyramid Medical Transformer (ours)** | **$0.81\pm0.05$** - | **$0.80\pm0.07$** - | **$0.80\pm0.06$** - |

Table 1. Model comparison of our Pyramid Medical Transformer against other CNN-based and full attention-based methods. We compare the Dice Similarity Coefficient in percentage for both datasets across seven models. The second row for each method reports the p-value of comparing to our proposed method, in which the number of images are 80, 896, and 5904 for GLAS, MoNuSeg, and HECKTOR, respectively.

We studied the effects of the new network structure and an adaptive partitioning mechanism (Figure 5). We evaluated the performance of combining only one of the short, mid, and long-range attention with the image features extracted by a CNN branch. For comparison, we tested the partitioning scheme used by MedTrans [1], which means we partition the feature maps into patches and perform "global" axial attention on H/4×W/4 patches. We kept the rest structure of the model and training parameters. The major difference between the partitioning scheme and our new method is that each pixel attends to a patch centered on itself instead of a pre-cut patch.

The results showed that on the GLAS and HECKTOR dataset, shorter-range branches have lower performance than long-range branches, and the PMTrans model that comprises three branches has the best performance. The difference between the short-range branch and the long-range branch is 7.52%. This effect can also be seen on the MoNuSeg dataset, but it is less obvious. The difference between the short-range branch and the long-range branch is 1.44%. The effectiveness of the non-trivial partitioning scheme varies due to the different sizes of the segmentation object from the three datasets. The gland segmentation targets are considerably larger than the nucleus, and it is more likely that a $\frac{H}{4} \times \frac{W}{4}$ patch cannot contain one object, which means the model needs to learn dependencies across longer ranges. However, for the MoNuSeg dataset, each nucleus is much smaller than a patch, so the difference between shorter-range land longer-range branches is not as significant. Moreover, we showed that the adaptive partitioning scheme improved the model performance on both datasets.

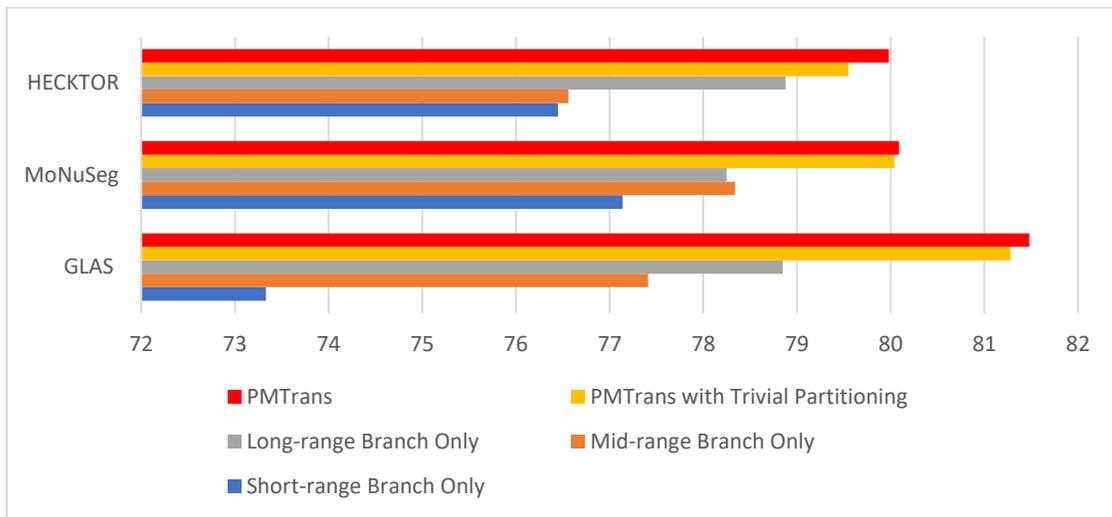

Fig.5 Ablation Study of Pyramid Structure and Adaptive Partitioning Mechanism

CNN-branch is integrated into the model to exploit the feature extraction power of convolutional layers. We tested three different CNN structures as well as removing the CNN branch (Table 3). The three different CNN backbones we tested are vanilla CNN [17], residual convolution blocks [34], and densely connected convolution blocks [19].

The results (Figure 6) showed that different CNN backbones have similar performance, and they all outperformed the model without a CNN branch. We performed this ablation study to show that the CNN branch boosts the model performance by adding feature extraction power of convolution layers. We chose to use residual connection backbones for the Pyramid Medical Transformer, considering the computational cost.

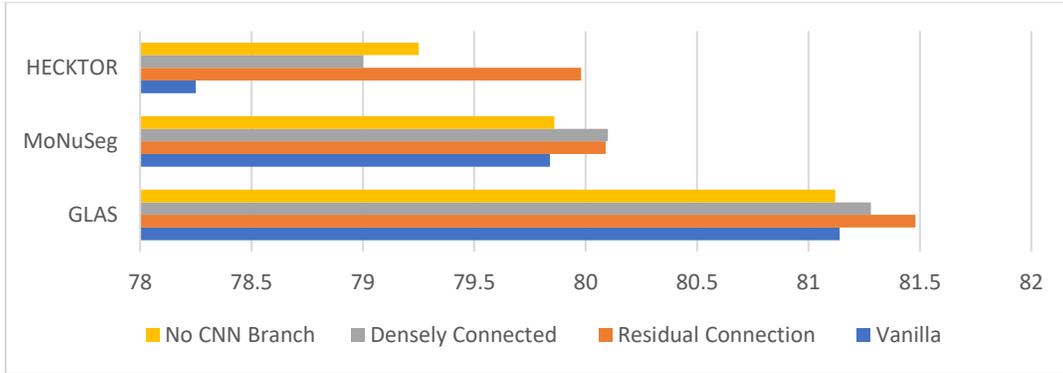

Fig.6 Ablation Study of Different CNN Branch

## 5 Discussion

Semantic segmentation is an essential task in medical imaging applications; accurate delineation of tumors enhances the outcome of diagnosis, treatment planning, and image-guided surgery. While CNN-based models capture long-range dependencies by inefficiently stacking convolution layers, attention-based models explicitly build relations of all ranges. However, assigning learnable parameters to all pairs of global relations is costly; thus, efficient approximations of global self-attention have been studied to circumvent the inherent quadratic complexity. The most popular technique is to partition the whole image into patches, which are subsequently used to compute (incomplete) global attention across patches [24] or pixel-wise global attention within patches [1]. This arbitrary partitioning mechanism imposes serious problems on existing medical segmentation methods since it cannot capture all attention of different scales and ranges. When large objects are partitioned into different patches, these fragments cannot effectively attend to each other. In this paper, we designed and developed a pyramid deep-learning architecture to address this critical issue.

The proposed method, Pyramid Medical Transformer (PMTrans), implements a multi-scale model architecture and exploits the feature extraction power of both self-attention and convolution layers. The new model uses multiple input resolutions to comprehend relations of different ranges without changing the overall complexity of self-attention computation. It learns longer-range dependencies on lower resolution images and refined segmentation details on full resolution images.

PMTrans also implements an adaptive partitioning mechanism to reserve informative relations that the existing rigid partitioning mechanism misses. These retained relations, which may be lost during the patch partitioning step in the existing methods, improve the segmentation outcome.

The results of PMTrans on three medical imaging datasets (GLAS [10], MoNuSeg [11], and HECKTOR [12]) showed its superior performance over the state-of-the-art convolution-based and transformer-based models on medical images. PMTrans is a general segmentation method designed for medical images but readily applicable to images from other domains.